\pdfoutput=1

\documentclass[11pt]{article}

\usepackage{emnlp2021}

\usepackage{times}
\usepackage{latexsym}

\usepackage{placeins}

\usepackage{tablefootnote}

\usepackage[T1]{fontenc}

\usepackage[utf8]{inputenc}

\usepackage{microtype}

\usepackage{booktabs}
\usepackage{amsmath}
\usepackage{graphicx}
\usepackage{multirow}

\expandafter\def\expandafter\UrlBreaks\expandafter{\UrlBreaks
  \do\a\do\b\do\c\do\d\do\e\do\f\do\g\do\h\do\i\do\j%
  \do\k\do\l\do\m\do\n\do\o\do\p\do\q\do\r\do\s\do\t%
  \do\u\do\v\do\w\do\x\do\y\do\z\do\A\do\B\do\C\do\D%
  \do\E\do\F\do\G\do\H\do\I\do\J\do\K\do\L\do\M\do\N%
  \do\O\do\P\do\Q\do\R\do\S\do\T\do\U\do\V\do\W\do\X%
  \do\Y\do\Z}

\usepackage{soulutf8}
\usepackage[normalem]{ulem}
\usepackage{xcolor}

\usepackage{enumitem}
\setlist{nolistsep}
\usepackage{microtype}

\iftrue 
    \usepackage[final]{proofing}
  \renewcommand\hl[1]{{#1}}  
   \newcommand\draftHL[2]{#1}{\draftnote{\red{#2}}}
   \newcommand\redHL[1]{}
  \newcommand\todo[1]{}

  \newcommand{\Djame}[1]{}

\else  

\usepackage[draft]{proofing}

\newcommand{\Djame}[1]{
\textbf{\textcolor{red}{\hl{Djame: #1}}}
}

\newcommand\red[1]{{{\textcolor{red}{\bf #1}}}}

\newcommand\draftHL[2]{\hl{#1}{\draftnote{\red{#2}}}}

\let\oldred\red
\renewcommand\red[1]{{ \oldred{{#1}}}}

 \renewcommand\draftHL[2]{\hl{#1}{\draftnote{\red{#2}}}}
 \newcommand\redHL[1]{\red{\hl{#1}}}
\let\olddraftnote\draftnote
\renewcommand\draftnote[1]{\olddraftnote{\red{#1}}}

\fi

\usepackage{tablefootnote} 

\title{PAGnol: An Extra-Large French Generative Model}

\author{
Julien Launay$^{* 1,2}$ \quad
Elena Tommasone$^{* 1}$ \quad 
Baptiste Pannier$^1$ \quad 
Fran\c{c}ois Boniface$^{\dag}$\\
\textbf{
 \quad
Am\'{e}lie Chatelain$^1$ \quad
Alessandro Cappelli$^1$ \quad
Iacopo Poli$^1$ \quad 
Djam\'{e} Seddah$^3$}
\\
$^1$ LightOn
$^2$ LPENS, École Normale Supérieure
$^3$ Inria, Paris
\\
\texttt{\{julien,elena,baptiste,amelie,alessandro,iacopo\}@lighton.ai} \\
\texttt{djame.seddah@inria.fr} 
\\
\\
\url{lair.lighton.ai/pagnol}
}

\begin{document}
\maketitle
\begin{abstract}
Access to large pre-trained models of varied architectures, in many different languages, is central to the democratization of NLP. We introduce PAGnol, a collection of French GPT models. Using scaling laws, we efficiently train PAGnol-XL (1.5B parameters) with the same computational budget as CamemBERT, a model 13 times smaller. PAGnol-XL is the largest model trained to date for the French language. We plan to train increasingly large and performing versions of PAGnol, exploring the capabilities of French extreme-scale models. 

For this first release, we focus on the pre-training and scaling calculations underlining PAGnol. We fit a scaling law for compute for the French language, and compare it with its English counterpart. We find the pre-training dataset significantly conditions the quality of the outputs, with common datasets such as OSCAR leading to low-quality offensive text. We evaluate our models on discriminative and generative tasks in French, comparing to other state-of-the-art French and multilingual models, and reaching the state of the art in the abstract summarization task. Our research was conducted on the public GENCI \textit{Jean Zay} supercomputer, and our models up to the Large are made publicly available. 
\end{abstract}

\renewcommand{\thefootnote}{\fnsymbol{footnote}}
\footnotetext[1]{Equal contribution.}
\footnotetext[2]{Work performed at LightOn.}
\renewcommand{\thefootnote}{\arabic{footnote}}

\section{Introduction}

Large pre-trained language models are the workhorses of modern Natural Language Processing (NLP). The use of scalable and efficient attention-based Transformers \cite{vaswani2017attention}, rather than recurrent neural networks, has enabled increasingly large and capable models. Through self-supervised learning, these models learn contextual word embeddings, building a general representation of language. After this \emph{pre-training} they can be \emph{fine-tuned} to target specific tasks (e.g. classification, parsing, summarization).

Three approaches dominate the field: (1) causal autoregressive decoder-only models, such as GPT \cite{gpt}, learning from a general language modelling tasks; (2) bidirectional encoder-only models, such as BERT \cite{bert}, learning from masked language modelling; (3) sequence-to-sequence models, such as BART \cite{bart} or T5 \cite{raffel2020t5}, combining both a bidirectional encoder and an autoregressive decoder, learning from a language denoising task. Encoder-only and sequence-to-sequence models excel in language understanding tasks, and have shadowed autoregressive models as a lesser option. 

Autoregressive models have been shown to predictably benefit from increased size \cite{kaplan2020scaling,henighan2020scaling}. Scaling laws establish a direct relationship between model size and end-task performance, justifying the training of increasingly large models \cite{gpt3,zeng2021pangu,kim2021changes,wei2021finetuned}. These laws can also inform design decisions, helping practitioners use their available compute budget optimally. A significant finding has been that larger models are more sample and compute efficient: with a given compute budget, it is preferable to train a larger model significantly short of convergence than to train a smaller model to convergence. Furthermore, at extreme-scale, such as the 175 billion parameters of GPT-3 \cite{gpt3}, autoregressive models exhibit unique few-shot abilities: they can learn from a few prompted examples, without weight updates. This capability questions the current fine-tuning paradigm, and may make billion parameters models more attractive and usable. Recent forays into \emph{prompt engineering/tuning} \cite{li2021prefix,lester2021power} have even seemingly bridged the gap between few-shot performance and fine-tuning.   

Encoder-only (CamemBERT \cite{camembert} and FlauBERT \cite{flaubert}) and sequence-to-sequence models (BARThez \cite{barthez}) exist for the French language, and recently a decoder-only model with 1 billions parameters has been made available \cite{simoulin2021modele}. We introduce PAGnol\footnote{PAG: \emph{Pré-Apprentissage Génératif}. Marcel Pagnol was a famous French novelist.} in this family, a collection of four French GPT-like models, and make the following contributions: 
\begin{itemize}
    \item \textbf{Largest French model.} We train on CCNet and publicly release four models, with up to 1.5B parameters for PAGnol-XL. At the time of this work, this is the largest non-sparse French language model available, and we plan to explore increasingly large and powerful models in the future. 
    \item \textbf{Optimal scaling.} We use scaling laws to inform our training setup, resulting in optimal use of our compute budget. PAGnol-XL is trained with a budget of only 3 PF-days, just as much as the 13 times smaller CamemBERT. From our collection of models, we adjust scaling laws for the French language.
    \item \textbf{Dataset suitability.} We highlight the importance of proper dataset pre-processing when training generative autoregressive models. While OSCAR has been relied on for French encoder-only models, we find it is not suited to PAGnol, leading to low-quality offensive outputs. 
    \item \textbf{End-task performance.} We evaluate on discriminative (FLUE) and generative tasks (question answering on FQuAD and summarization with OrangeSum) in the fine-tuning and prompt tuning regimes. We establish a new state of the art for summarization in French on OrangeSum.
\end{itemize}

\section{Related work} 
\paragraph{Language models.} The design and training of neural language models able to create and process word embeddings is the cornerstone of modern NLP. Early on, self-supervised learning was identified as an efficient and scalable way to train such models. The use of deeper and more complex neural architectures enabled going from static embeddings (\textit{word2vec} \cite{mikolov2013efficient}, \textit{GloVe} \cite{pennington2014glove}) to contextual embeddings, allowing models to deal with polysemy. Although approaches such as ELMo \cite{peters2018deep} and ULMFiT \cite{howard2018universal} highlighted that learned representations can be transferred across downstream tasks, the poor scalability of RNNs prevented this vision from being fully realized.

By getting rid of the costly and delicate recurrent processing, attention-based Transformers \cite{vaswani2017attention} spurred a wide interest in NLP. GPT \cite{gpt}, a decoder-only variant of Transformers, demonstrated large-scale transfer learning from general language modelling to 12 NLU tasks. Along with the rise of easy-to-use libraries, encoder-only BERT \cite{bert}, relying on masked language modeling, made NLP a commodity -- wherein every practitioner could rely on a pre-trained language model and fine-tune it cheaply to a task of interest. BERT models are limited by their ability to only "fill-in-the-gap" for a span of words: this forbids their use in generative tasks (e.g. summarization). 

With sequence to sequence models and pre-training through denoising tasks, the original architecture of Transformers made a comeback with BART \cite{bart}, bridging the gap between the generative capabilities of decoder-only models and the downstream task performance of encoder-only models. Through gradually larger and more powerful architectures, state-of-the-art models are approaching human-level performance on many tasks.

Successive generations of GPT models have questioned the current fine-tuning paradigm. GPT-2 \cite{gpt2}, with 1.5 billion parameters, demonstrated that large language models could tackle entirely new tasks through \emph{few-shot learning}\footnote{In other areas of machine learning, this has been referred to as \emph{zero-shot learning}, as no weight updates are necessary.}. Without any fine-tuning, from just a few prompted examples, GPT-2 achieved fair performance on a number of complex downstream tasks. Furthering this endeavour, GPT-3 \cite{gpt3}, with 175 billion parameters, achieved state-of-the-art performance on some tasks, without the need for fine-tuning. This opens new possibilities for low-resources tasks, as well as paths to more natural interactions with these models: recent research suggests the gap between few-shot learning and fine-tuning may even be bridged through so-called prompt programming/tuning \cite{li2021prefix,lester2021power}. 

\paragraph{Scaling laws.} More specifically to our setting, neural language models have been shown to predictably benefit from increased scale \cite{kaplan2020scaling}. Their training dynamics are size-invariant, allowing test loss, parameter count, and dataset size to be correlated through smooth scaling laws. This is in fact true of all GPT-like autoregressive models, even when applied to image, multimodal, or mathematics modeling \cite{henighan2020scaling}. Gains in autoregressive cross-entropy loss also directly translates to gains in end-task performance after fine-tuning. As they relate to compute budget, these predictions can be used to inform the training of large models.

\paragraph{Non-English generative models.} BERT-like models are now available in a broad number of languages, either as specialized models are as multilingual ones. This is less so the case for generative models, perhaps because of issues in controlling the language used at generation time. For the French language, GPT\textsubscript{fr} is an autoregressive generative model, and BARThez \cite{barthez} targets some generative abilities. Smaller-scale efforts exist, such as BelGPT \cite{louis2020belgpt2}, but they are limited to small models. GPT models have been trained for German \cite{stefan_schweter_2020_4275046}, Chinese \cite{zeng2021pangu}, and the Arabic language \cite{antoun-etal-2021-aragpt2}, among others.

\section{Efficient training with scaling laws}

\paragraph{Scaling.} We use scaling laws to inform the duration of the training of our largest models. Rather than training to convergence, which would be wasteful, we train to optimality, as predicted by the equations provided in \cite{kaplan2020scaling}. This is akin to what has been done for GPT-3, and this enables us to keep our computational budget in line with that of CamemBERT, a model 13x smaller than PAGnol-XL. We find that training all of our models for a single epoch on the 30GT of CCNet enables us to reach optimality for the most expensive XL model. Table \ref{tab:architectures} presents the ratios between the compute budget effectively used and that to optimality ($r^C_\text{opt}$) or to convergence ($r^C_\text{conv}$). While our small model is trained to convergence,  others are trained significantly short of it. We find that our training performance matches nicely with the estimated $2.6$ PF-days for the training of GPT\textsubscript{fr}-LARGE from \cite{simoulin2021modele}. 

\section{PAGnol}
In this section, we describe the data, model, and training specifications for PAGnol. In Table \ref{tab:comparison}, we highlight some of the key differences and similarities with other French models, and in Table \ref{tab:multilingual} we present two multilingual models that we consider in the following.

\begin{table*}[t]
\begin{minipage}{\linewidth}
\centering

\scalebox{0.7}{\begin{tabular}{@{}lcccc|c@{}}
\toprule
                                  & \textbf{CamemBERT} & \textbf{FlauBERT} & \textbf{BARThez}  & \textbf{GPT\textsubscript{fr}}                                                             & \textbf{PAGnol (ours)} \\ \midrule
\textbf{Language}                 & French             & French          & French            & French                                                               & French                 \\
\textbf{Parameters}               & 110/335M           & 138/373M            & 165M              & 124M/1B                                                                       & 124M/355M/773M/1.5B    \\
\textbf{Context}               & 512           & 512          & 768              & 1024                                                                       & 1024/2048    \\
\multirow{2}{*}{\textbf{Dataset}} & OSCAR              & Custom\footnote{\label{fnote:data}FlauBERT and BARThez use a similar pre-training dataset assembling CommonCrawl, NewsCrawl, Wikipedia, and other smaller corpora.}       & Custom$^\text{\ref{fnote:data}}$           & Filtered Common Crawl                                                                       & CCNet                  \\
                                  & 33GT/138GB         & 13GT/71GB                & 66GB              & 1.6/3.11 GT & 32GT                   \\
\textbf{Tokenization}             & SentencePiece 32k  & BPE 50k & SentencePiece 50k & BPE 50k                                                         & BPE 50k                \\
\textbf{Compute [PF-days]}        & 3/10               & $\sim7/26$\footnote{\label{fnote:compute}Insufficient data was provided by authors to infer compute budgets properly.}            & $\sim4^\text{\ref{fnote:compute}}$    & ?/2.6                                                           & 0.3/0.7/2/3            \\ \bottomrule
\end{tabular}}

\caption{Model, data, and training setup for PAGnol and other French models. Data size is reported in gigatokens (GT), and compute in PF-days ($8.64 \times 10^{19}$ FLOP). PAGnol is the largest French model. Despite being significantly larger than existing models, its compute cost remains reasonable: as recommended by scaling laws, we train models to optimality, and not to convergence.}
\label{tab:comparison}
\end{minipage}
\end{table*}

\begin{table}[t]
\centering
\scalebox{0.7}{\begin{tabular}{@{}lcc@{}}
\toprule
                  & \textbf{mBERT} & \textbf{mBART} \\ \midrule
\textbf{Language} & 104 languages & 25 languages \\
\textbf{Parameters} & 110M        & 610M \\
\textbf{Context} & 512           & 768   \\
\multirow{2}{*}{\textbf{Dataset}} & Wikipedia & CC25\\
                                 &            & \begin{tabular}[x]{@{}c@{}}180GT/1369GB\\(10GT/57GB French)\end{tabular}  \\
\textbf{Tokenization}             & WordPiece 110k  & SentencePiece 250k \\
\textbf{Compute [PF-days]} & 2  & $\sim60^\text{\ref{fnote:compute}}$ \\ \bottomrule
\end{tabular}}
\caption{Model, data, and training setup for multilingual models including French that we consider. Data size is reported in gigatokens (GT), and compute in PF-days ($8.64 \times 10^{19}$ FLOP).}
\label{tab:multilingual}
\end{table}

\subsection{Pre-training data}

\paragraph{Sourcing.} The Common Crawl (CC) project browses and indexes all content available online. It generates 200-300 TiB of data per month (around 5\% of which is in French), and constitutes the bulk of most NLP datasets nowadays. We consider in our experiments two datasets based on CommonCrawl data: CCNet \cite{wenzek2020ccnet} and OSCAR \cite{oscar}. Once tokenized, OSCAR contains 33GT and CCNet 32GT. We use CCNet for all our main experiments and released models, and compare with results obtained on OSCAR in Section \ref{sec:oscar}. We validate on the fr-wiki dataset (0.5GT) and French TreeBank (650kT) \cite{abeille2003building}.

\paragraph{CCNet.} CCNet combines the usual fastText \cite{joulin2017bag} pre-processing of CC data with an additional filtering step to select high-quality documents. This filtering is done through a language model trained on Wikipedia, ensuring a text quality similar to that of its articles. We use a version of CCNet identical to the one considered in the CamemBERT paper. 

\paragraph{OSCAR.} OSCAR uses a fastText classifier to select documents and identify their languages, without any additional filtering. OSCAR is thus more varied, but more "noisy", than CCNet. OSCAR has been used to train other French language models such as CamemBERT. 

\paragraph{Tokenization.} We use byte-level Byte-Pair Encoding (BPE), with a vocabulary of 50,262 tokens: 256 bytes, 6 special tokens, and 50,000 merges. Paragraphs are separated by an <EOS> token and documents are separated by a <SEP> token. We add a prefix space before tokenization, so that the first word in a paragraph is tokenized in the same way as if it was at any other position. This is similar to the setup of FlauBERT and GPT-2. For the models trained on OSCAR, we use a slightly smaller vocabulary size of 50,000 tokens.

\subsection{Model specification}
\begin{table}[h]
\centering

\scalebox{0.7}{\begin{tabular}{@{}cccccccc@{}}
\toprule
\textbf{PAGnol}      & $n_\text{params}$ & $n_\text{layers}$ & $d_\text{model}$ & $n_\text{heads}$ & $C$ [PF-days] & $r^C_\text{conv}$ & $r^C_\text{opt}$ \\ \midrule
\textbf{S}  & 124M              & 12                & 768              & 12               & 0.3           & 1,3              & 9,0               \\
\textbf{M}  & 355M              & 24                & 1024             & 16               & 0.7           & 0,5              & 3,0               \\
\textbf{L}  & 773M              & 36                & 1280             & 20               & 2             & 0,4              & 2,5               \\
\textbf{XL} & 1.5B              & 48                & 1600             & 25               & 3             & 0,2              & 1,3               \\ \bottomrule
\end{tabular}}
\caption{Model and training budgets for PAGnol. All models are trained on a single epoch of our 32GT CCNet-curated data. $C$ is the compute budget used for the training of the model. $r^C_\text{opt}$ is the ratio between $C$ and $C_\text{opt}$, the optimal compute budget derived from scaling laws. $r^C_\text{conv}$ is the ratio between $C$ and $C_\text{conv}$, the compute budget derived from scaling laws to train the model to convergence.}
\label{tab:architectures}
\end{table}

PAGnol is a decoder-only autoregressive transformer, \draftHL{reproducing the architectural choices of GPT-3, with up to 1.5 billions parameters.}{Shouldn't this be GPT3? which parts of the models are from gpt2 and which parts exactly are from gpt3? -ds  Does this look ok now? I also added a part on rotary embeddings at the end of this paragraph. The equations are a bit heavy, not sure we want to include them -ip}. We evaluate four model sizes: small, medium, large, and extra-large, with architectures detailed in Table \ref{tab:architectures}. We use a context size of 1,024 tokens for the S, M and L models. The XL uses a context size of 2,048, the largest at release for a French model. Additionally, we use Rotary Embeddings \cite{su2021roformer} in place of Learned Positional Embeddings for the XL model, since they provide much better training stability at the billion parameters regime.

\subsection{Pre-training} 

\paragraph{Training objective.} We use an autoregressive language modelling objective, where the model learns to predict the next word in a sentence. To improve efficiency, we always fill the context with as much text at possible, and inform the model about separate documents through the <SEP> token. 

\paragraph{Optimization.} We use the Adam optimizer \cite{kingma2014adam} with a warmup followed by cosine decay learning rate schedule. We find that proper initialization is key to training stability, and reproduce the setup effectively implemented by Megatron-LM \cite{shoeybi2019megatron}. We initialize all weights with a normal distribution $\mathcal{N}(0, 0.02)$ and scale the weights under the residual layers by $1 / \sqrt{2n_\text{layers}}$. We tune hyperparameters over 10k step first, and pick the set with the best train perplexity.

\paragraph{Distributed training.} All training runs were performed on the public GENCI supercomputer \emph{Jean Zay}, on nodes with 4x or 8x V100 32GB and a 100Gb/s interconnect. We built our own GPT implementation from scratch in PyTorch \cite{paszke2019pytorch}, leveraging FairScale for distributed training \cite{FairScale2021}.

Models up to PAGnol-L can be trained using simple distributed data parallelism (DDP). However, PAGnol-XL does not fit in 32GB of memory. We use optimizer and model state sharding, along with activation checkpointing and CPU offloading to fit the model in memory. This results in a setup similar to ZeRO-3 \cite{rajbhandari2021zero}. It is beneficial to train even small models with this setup, as it allows for a larger batch size, and significantly higher GPU throughput. 

\begin{table}[h]
\centering
\scalebox{0.7}{\begin{tabular}{@{}cccccccc@{}}
\toprule
\textbf{PAGnol}      & $n_\text{params}$ & fr-wiki & FTB: whole~~~ (train/val/test)\\ \midrule
\textbf{S}  & 124M              & 43.38 & 23.87~~~ (23.90, 24.38, 23.34)                     \\
\textbf{M}  & 355M              & 37.46 & 20.90~~~ (20.92, 21.59, 20.46)                       \\
\textbf{L}  & 773M              & 34.81 & 19.97~~~ (19.88, 21.19, 20.02)                        \\
\textbf{XL} & 1.5B              & 28.85 & 16.18~~~ (16.11, 16.67, 16.40)                    \\ \bottomrule
\end{tabular}}
\caption{Validation perplexity on fr-wiki and on the whole French TreeBank (FTB) for PAGnol models after 1 epoch of training on 30GT of CCNet.}
\label{tab:model_perp}
\end{table}

\paragraph{Perplexity.} We report final validation perplexity after 1 epoch over 30GT in Table \ref{tab:model_perp}. We use the the official
2019 French Wikipedia dumps and the French TreeBank dataset \cite{abeille2003building} in its SPMRL instance \cite{seddah-etal-2013-overview} as our validation sets. Because we are running all models for a single epoch on our data, there are limited risks of overfitting and memorization.

\paragraph{Scaling law for PAGnol models}
We fit a scaling law with the same functional form of \cite{kaplan2020scaling}, that is the following power law:
\begin{equation*}
    \mathcal{L} = \left ( \frac{k}{C} \right )^\alpha
\end{equation*}
where $\mathcal{L}$ is the validation loss of PAGnol models trained on CCNet, $k$ is a constant, $C$ is the compute in PF-days, and $\alpha$ is the scaling exponent. The fit is performed in log-log, and constrained to remain under the efficient frontier, using \textit{cvxpy} \cite{agrawal2018rewriting}. We exclude the L and XL models from the fit: due to the HPC environment, the training was performed in multiple splits. At restart, the optimizer state is not necessarily available, generating artefacts in the training and validation curves. Additionally, the use of Rotary Embeddings for the XL model would affect the scaling, and make it incomparable with the English models. We therefore trained two smaller models, an XXS and an XS, following the same architectural decisions of the larger ones, on the same datasets, and used these to fit a scaling law. We find a scaling exponent $\alpha=-0.036$ for the French language, to compare to the $-0.050$ for the English language from \cite{kaplan2020scaling}. With the relatively important caveats that we are using different datasets, codebase, and hardware, it appears that French is less compute efficient than English, and that for a same improvement in validation loss, therefore we need to spend more compute for French than for English. The increase morphological complexity of French  when compared to English \cite{seddah-etal-2010-lemmatization} and its, in average, longer sentences  could be a factor explaining this discrepancy.
\begin{figure}
    \centering
    \includegraphics[width=0.9\linewidth]{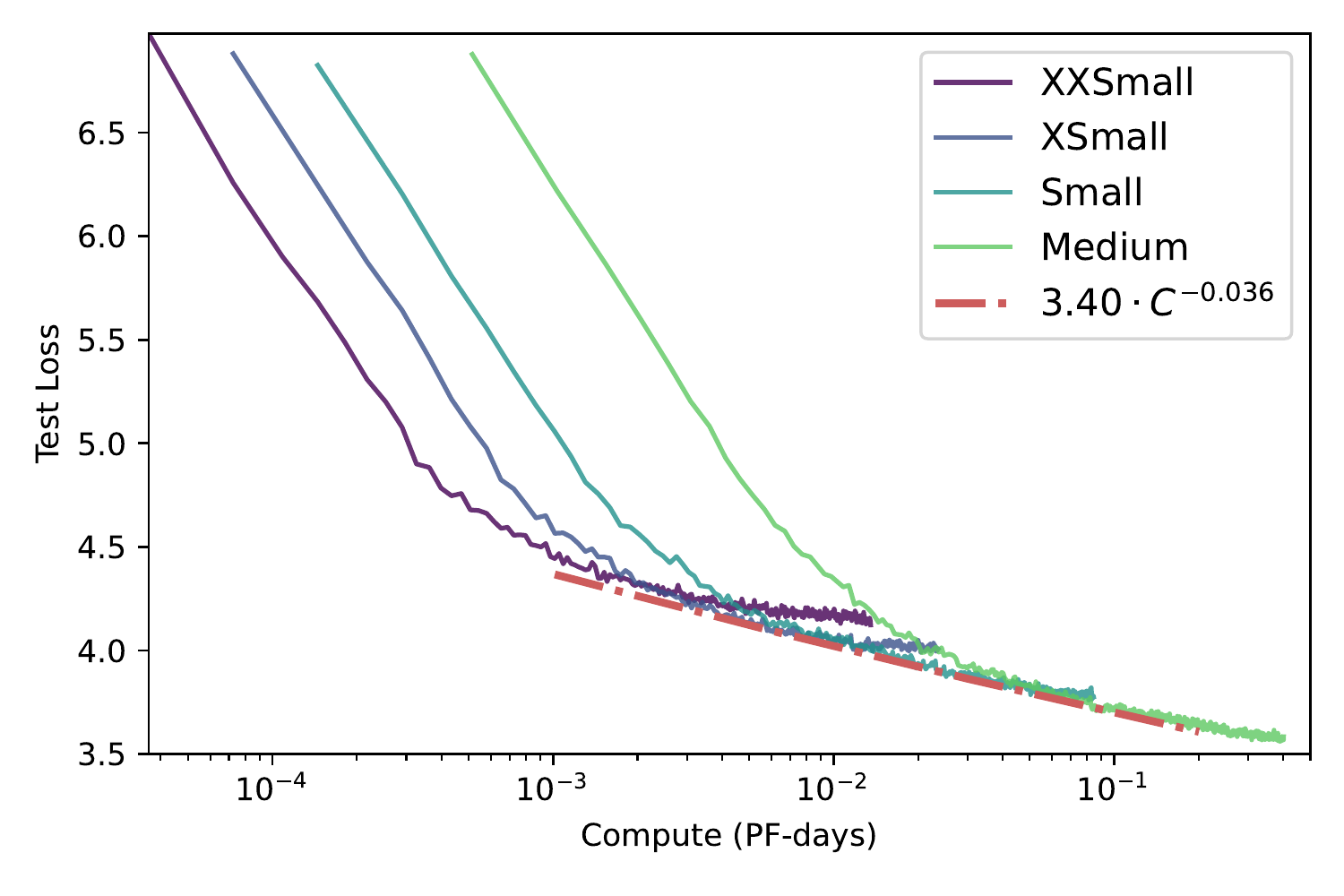}
    \caption{Scaling law relative to the compute for PAGnol models from XXS to M. We do not include L and XL: the interrupted nature of the training due to the HPC environment and the choice of Rotary Embeddings for the XL pollute validation curves with artefacts that negatively affect the quality of the fit.}
    \label{fig:scaling_law}
\end{figure}

\section{Influence of the pre-training data}
\label{sec:oscar}

Existing French models (CamemBERT, FlauBERT, BARThez) have been trained on datasets based on a simple filtering pipeline. A fastText classifier is used to isolate content in French, deduplication is applied, and noisy content (phone numbers, code, etc.) is removed. While this has been sufficient for pre-training encoder-only models and sequence-to-sequence models, the lack of quality control may be an issue for free-form generation with a decoder-only model such as PAGnol. Moreover, recent work on the OSCAR dataset (used by CamemBERT) has found that it may contain up to 5\% of non-linguistic content, or content in the wrong language, and 0.5\% of explicit content for the French language \cite{caswell2021quality}.

We initially pre-trained PAGnol on the OSCAR dataset, and while experimenting with the model, we observed the model generated offensive and explicit content, even when not prompted for it. For instance, the prompt \emph{Bonjour je suis} (Hello I am) often resulted in pornographic content, despite being rather naive. This motivated our choice to switch to CCNet instead. For research purposes, we release the small and medium models trained on OSCAR. In future iterations of this document, we will provide a more detailed investigation of the content generated by PAGnol-CCNet and by PAGnol-OSCAR.

\section{End-task performance}\label{sec:downstream}

\subsection{Discriminative tasks: FLUE}
We evaluate our models on the Sentence Classification task of the FLUE evaluation setup \cite{flaubert}. The task is a binary classification problem on reviews of Books, Music and DVD taken from the Amazon website. Each review is assigned a score from 1 to 5, and then labeled as "negative" if the score is lower than 3 and "positive" otherwise. We also evaluate our models on the paraphrasing and natural language inference tasks (PAWS-X and XNLI). PAWS-X consists in a binary classification task where the model has to identify whether two sentences are semantically equivalent or not. XNLI is instead a 3 class problem where we have to determine if a premise contradicts, entails or neither a given hypothesis. For the training, we add a CLS token at the end of the review (but before the EOS token). We then replace the projector at the end of the model with a linear layer and use the embedding of the CLS token to perform the classification. \newline
Table \ref{tab:cls} reports the test accuracy of the best
\draftHL{hyperparameter configuration}{Should we write the
  hyperparameters of the best runs as well? I think the table might
  become a bit cumbersome -ip -ds Maybe in the appendix ?} along with a comparison with other French language models. All models are fine-tuned for 6 epochs, except the medium OSCAR and the extra-large CC-100 which were trained respectively for 4 and 5 epochs. For each model, we finetune the learning rate and weight decay in the interval $[10^{-6}, 10^{-4}]$ and $[0, 10^{-3}]$ respectively. For the classification task, we use a cosine annealing scheduler that decays down to $1/10$ of the original learning rate in 5 epochs (3 for the medium OSCAR and 4 for the extra-large CC-100). We additionally checked if adding dropout with $p=0.1$ could improve the performance. For the PAWS-X and XNLI tasks, we finetune the learning rate in the interval $[10^{-6}, 10^{-4}]$. We use the cosine annealing scheduler down to a learning rate equal to $1/10$ of the original value ($1/5$ for the Small models) over $9/40$ Million tokens respectively. PAWS-X training is over 2 epochs while XNLI training over 1. PAGnol models slightly underperform smaller BERT models, while being better than multilingual alternatives, and their GPT\textsubscript{fr} counterparts. For PAGnol, performance improves with size but seems to saturate with the XL model, possibly because we had to use a lower batch size to fit on the hardware for fine-tuning. Additionally, while the generation quality of models trained on OSCAR is noticeably worse, they perform as well or better than the corresponding models trained on CCNet on these discriminative tasks.\newline

\begin{table*}[h]
\centering
\scalebox{0.8}{\begin{tabular}{ccccccc}
\toprule
Model           & Parameters & Books & Music & DVD & PAWS-X & XNLI \\ 
\midrule
MultiFiT							& Not Specified & 91.25 & 89.55 & 93.40 & - & - \\

mBERT								& 110/340 M*& 86.15 & 86.90 & 86.65 & 89.30 & 76.9 \\
mBART								& 610 M 	& 93.40 & 93.13 & 93.10 & 89.70 & 81.07\\
BARThez								& 216 M 	& 94.47 & 94.97 & 93.17 & 88.90 & 80.73\\
CamemBERT-BASE		& 110 M 	& 92.30 & 94.85 & 93.00 & 90.14 & 81.20\\
CamemBERT-LARGE		& 335 M 	& \textbf{95.47} & \textbf{96.00} & 95.37 & \textbf{91.83} & \textbf{85.33}\\
Flaubert-BASE 		& 138 M & 93.10 & 92.45 & 94.10 & 89.49 & 80.60\\ 
Flaubert-LARGE 		& 373 M & 95.00 & 94.10 & \textbf{95.85} & 89.34 & 83.40\\ 
\midrule
GPT\textsubscript{fr}-BASE  & 124 M & 88.30 & 86.90 & 89.30 & 83.30 & 75.60 \\
GPT\textsubscript{fr}-LARGE & 1 B & 91.60 & 91.40 & 92.60 & 86.30 & 77.90 \\
\midrule
PAGnol-S \textsubscript{OSCAR}     	& 124 M & 92.05 & 92.60 & 91.70 & 84.19 & 76.10 \\ 
PAGnol-M \textsubscript{OSCAR}     	& 355 M & 94.40 & 94.90 & \underline{94.30} & 87.44 & 79.46 \\ 
\midrule
PAGnol-S \textsubscript{CC-100}     	& 124 M & 92.00 & 93.00 & 91.65 & 87.19 & 75.67 \\ 
PAGnol-M \textsubscript{CC-100}     	& 355 M & 94.40 & 95.20 & 93.70 & 89.14 & 79.00 \\ 
PAGnol-L \textsubscript{CC-100}     	& 773 M & \underline{94.65} & 95.25 & 94.00 & \underline{90.70} & 81.48 \\ 
PAGnol-XL \textsubscript{CC-100}     	& 1.5 B & \underline{94.65} & \underline{95.35} & 94.18 & 89.47 & \underline{81.83} \\ 
\bottomrule
\end{tabular}}
\caption{Results on the FLUE Benchmark including classification (Books, Music, DVD), paraphrasing (PAWS-X) and natural language inference (XNLI) tasks. The best overall results are highlighted in \textbf{bold}, and the best results for GPT models are \underline{underlined}.}
\label{tab:cls}
\end{table*}

\subsection{Generative task: FQuAD}
FQuAD \cite{d2020fquad} is a native French question answering dataset, comprising more than 25.000 questions fabricated by higher education students from a set of Wikipedia articles, following the same philosophy as the English dataset SQuAD \cite{rajpurkar2018know}. Given a document $d_{i}$, a question $q_{i}$ and the corresponding answer $a_{i}$, the Question Answering task is casted into this format: 
\begin{equation*}
\mbox{\textit{"$\{d_{i}\}$ Question: $\{q_{i}\}$ Réponse: $\{a_{i}\}$"}}
\end{equation*}
where \textit{Réponse} corresponds to \textit{Answer} in French.\\\\
Given this input format, in a setup similar to pretraining, the likelihood of the sequence corresponding to the answer is maximized using the cross entropy loss on the tokens corresponding to the answer. We use the Adam optimizer and finetune the learning rate and weight decay in the interval $[10^{-6}, 10^{-4}]$ and $[0, 10^{-3}]$. The different models were trained for 2 epochs. As noted by \newcite{gpt2}, the performance of autoregressive models is still worse than question answering systems based on masked language models. Indeed, we evaluated the finetuning of OpenAI GPT-2 small and medium on SQuAD, and obtained EM and F1 scores in the same range of PAGnol on FQuAD (Table~\ref{tab:gpt:squad}).

\begin{table}[h]
\centering
\scalebox{0.7}{\begin{tabular}{ccc}
\toprule
Model & EM & F1 \\ 
\midrule
CamemBERT-LARGE & \textbf{82.1} & \textbf{92.2} \\
CamemBERT-BASE & 78.4	& 88.4 \\
\midrule
PAGnol-S \textsubscript{OSCAR} & 31.7 & 52.8 \\
PAGnol-M \textsubscript{OSCAR} & 37.1 & 59.4 \\
\midrule
PAGnol-S \textsubscript{CC-100} & 33.7 & 56.0 \\
PAGnol-M \textsubscript{CC-100} & 36.8 & 59.0 \\
PAGnol-L \textsubscript{CC-100} & 42.8 & 66.3 \\
PAGnol-XL \textsubscript{CC-100} & 44.4 & 68.5 \\
\bottomrule
\end{tabular}}
\caption{Question answering on FQuAD.}
\label{tab:fquad}
\end{table}

\begin{table}[h!]
\begin{center}
{\footnotesize
\begin{tabular}{rccc}
\hline
{ Model} & { Size} & { EM} & { F1} \\
\hline
GPT & small & 45.5 & 62.7 \\
GPT & medium & 50.8 & 68.1 \\
\hline
\end{tabular}
\caption{GPT-2 small and medium model performance on SQuaD}
\label{tab:gpt:squad}
}
\end{center}
\end{table}


\subsection{Generative task: OrangeSum}
OrangeSum \cite{eddine2020barthez} is a summarization dataset, considered to be the French equivalent of the XSum \cite{xsum-emnlp}. It is an abstractive dataset containing summary of news article from the "Orange Actu" website. Each article comes with a professionally-written title and abstract. Hence, the dataset includes two tasks: OrangeSum Title and OrangeSum Abstract. We evaluate PAGnol on the latter. 

Similarly to our setup for question answering, given a news article $n_{i}$ and an abstract $a_{i}$, we cast the summarization task in this format:
\begin{equation*}
\mbox{\textit{"$\{n_{i}\}$ Summary: $\{a_{i}\}$"}} 
\end{equation*}
We finetune our model on the crossentropy loss computed only on the tokens of the produced summary. We optimize the learning rate and weight decay in the same interval as FLUE, using the same scheduler, and train for $4$ epochs. We add a dropout with $p=0.1$ to improve the performance. We evaluate the fine-tuned model using greedy token generation and the ROUGE metric. This task, more geared towards generation, sees PAGnol-XL establish a new state of the art for summarization on OrangeSum. 

\begin{table}[h]
\centering
\scalebox{0.7}{\begin{tabular}{ccccc}
\toprule
Model           & Parameters & R-1 & R-2 & R-L   \\ 
\midrule
BARThez								& 216 M 	& \textbf{31.44} & \underline{12.77} & \underline{22.23} \\
\midrule
PAGnol-S \textsubscript{OSCAR}     	& 124 M & 22.79 & 6.16 & 16.03 \\ 
PAGnol-M \textsubscript{OSCAR}     	& 355 M & 24.89 & 7.87 & 17.78 \\ 
\midrule
PAGnol-S \textsubscript{CC-100}     	& 124 M & 26.47 & 9.49 & 17.12 \\ 
PAGnol-M \textsubscript{CC-100}     	& 355 M & 28.20 & 10.80 & 20.79\\ 
PAGnol-L \textsubscript{CC-100}     	& 773 M & 28.12 & 11.05 & 20.81 \\ 
PAGnol-XL \textsubscript{CC-100}     	& 1.5 B & \underline{31.17} & \textbf{12.86} & \textbf{22.50} \\ 
\bottomrule
\end{tabular}}
\caption{Text summarization on the OrangeSum Abstract task. Best results are highlighted in \textbf{bold}, and second best are \underline{underlined}.}
\label{tab:sum}
\end{table}

\section{Prompt Tuning}
Human prompt engineering to extract good zero- and few-shot performance for large language models has motivated research in \textit{prompt tuning}: placing some random vectors in the input sequence and optimizing their values, while keeping the pre-trained model weights fixed. The advantage of this approach is that the model does not change, and only the prompt is optimized. We follow the approach in \cite{lester2021power}, and optimize a certain number $k$ of tokens in our soft prompt for the three aforementioned tasks. The best hyperparameters per size per task have been selected through a grid search for the value of $k$, learning rate and dropout. In particular we performed a grid search over $k=\{1, 5, 20, 35, 50\}$, learning rate in $\{0.3, 0.1, 0.01, 0.001, 0.0005\}$, and dropout values in $\{0, 0.01, 0.1\}$. We show the results for FLUE, FQuAD, and OrangeSum in Tables \ref{tab:soft_cls}, \ref{tab:soft_fquad} and \ref{tab:soft_sum}. We expected a smooth scaling in performance with size and to progressively close the gap with fine-tuning performance, as shown by \cite{lester2021power}, however this scaling slows significantly when we reach the XL model. We suspect a bug in our implementation of prompt tuning with Rotary Embeddings, causing the performance hit, therefore we temporarily show the results for the XL model in \textit{italic} in this setting. This is a work in progress and will be updated in a new release.

\begin{table}[h]
\centering
\scalebox{0.7}{\begin{tabular}{ccccc}
\toprule
PAGnol & Books & Music & DVD   \\ 
\midrule
S & 88.50 & 87.95 & 88.24 \\ 
M & 91.60 & 92.65 & 90.69 \\ 
L & 92.60 & 93.1 & 91.69 \\ 
XL & \textit{92.50} & \textit{93.25} & \textit{92.14} \\ 
\bottomrule
\end{tabular}}
\caption{Prompt Tuning performance for sentence classification (CLS).}
\label{tab:soft_cls}
\end{table}

\begin{table}[h]
\centering
\scalebox{0.7}{\begin{tabular}{cccc}
\toprule
PAGnol & EM & F1 \\ 
\midrule
S & 0.243 & 0.427 \\
M & 0.320 & 0.561 \\
L & 0.365 & 0.526 \\
XL & \textit{0.403} & \textit{0.450} \\
\bottomrule
\end{tabular}}
\caption{Prompt Tuning performance for question answering on fQuad.}
\label{tab:soft_fquad}
\end{table}

\begin{table}[h]
\centering
\scalebox{0.7}{\begin{tabular}{cccc}
\toprule
PAGnol & R-1 & R-2 & R-L   \\ 
\midrule
S & 24.54 & 8.98 & 18.45 \\ 
M & 27.80 & 10.56 & 20.29\\ 
L & 28.25 & 11.05 & 21.03 \\ 
XL & \textit{28.72} & \textit{11.08} & \textit{20.89} \\ 
\bottomrule
\end{tabular}}
\caption{Prompt tuning performance for text summarization on OrangeSum.}
\label{tab:soft_sum}
\end{table}


\section{Discussion}
Without fear of aligning an overused {\em clich\'e}, the release of
large language neural models have not only revolutionized the NLP
field by bringing a major leap in performance in almost every tasks
they were applied to, they crucially changed the perception of the
risks of their potential misuse. The point is that this dramatic boost
of performance has led the field to rely on the capacity of those
large models to transfer their, in layman's terms, ``knowledge'' to
other tasks via various transfer learning modalities. Yet, with this
transfer, the potential data biases inherent to large corpus
collection used for pre-training are also susceptible to appear.
\newcite{gehman-etal-2020-realtoxicityprompts} thoroughly demonstrated
that all generative language models they tested (from GPT1 \cite{gpt}
trained on Book Corpus only to GPT3 \cite{gpt3} and CTRL
\cite{Keskar2019CTRLAC} trained on various corpora, including
user-generated content and web-crawled data sets) were capable of
producing toxic output in specific conditions\iftrue and presented
different ways of alleviating this behaviour\fi. Having been
pre-trained on Common Crawl-based corpora, our models are certainly
not immune from toxic content generation. 

\newcite{gehman-etal-2020-realtoxicityprompts} explored
two main ways to filter toxic outputs, one being based on the pursuing the pretraining on
less toxic data sets or on toxicity-annotated data set; the other
focusing on  the alteration of the decoding
strategy. \newcite{Krause2020GeDiGD} proposed to guide the output with
smaller language models. Given the interest in generative models, this
is an active line of research that we are currently pursuing.

More generally, the question of knowing whether the
pre-training data should be curated more or should the model output,
depending on the downstream application in sight, be debiased, or filtered, directly  is still
the object of vivid debates among the community
\cite{Bender_et_al:2021:Parrots,goldberg:2021:reparrots}, while of
course there is an agreement toward responsible use of such
technology. 

In this aspect, the release of a GPT-generated text detector by
\newcite{Antoun2021AraGPT2PT} along their Arabic GPT2 model is an interesting
step toward this direction. 
 
Regarding the environmental aspects of this model, our model pretraining
experiments consumed about 62k gpu hours from the Jean Zay HPC cluster. Being
based in France, its energy mix is made of nuclear (65-75\%), 20\%
renewable and the remaining with gas (or more rarely coal when
imported from abroad) (S.Requena, Dir. of Jean Zay, P.C).     

Regarding the performance of our models which almost constantly
outperform their closest French counterparts, the
GPT\textsubscript{fr} models \cite{simoulin2021modele}, one explaining
factor could be the size of our pretraining data set (30B token vs
3B). Given our computing limitation, we choose from the beginning to
use an experimental protocol as comparable as possible to the one used
for the CamemBERT model evaluation \cite{camembert}, it would of
course be interesting to perform a head to head comparison with the
GPT\textsubscript{fr} model. In terms of raw performance, it has been
constantly reported that GPT-based models on the billion parameter
scale provide inferior performance when compared to their ``regular''
transformer counterparts in classic fine-tuning scenarios, our results
confirm this for French as well while highlighting the interest of our
models in generation-oriented scenarios (such as text summarization
where PagnolXL establishes a new state of the art for French). As for
English \cite{lester2021power}, our encouraging preliminary
prompt-tuning results suggest that this approach is promising and
could be a way to close this performance gap.

\section{Conclusion}
We presented the Pagnol model collection, the first released large
scale generative model for French\footnote{Released on May 4th, 2021.
  \url{https://twitter.com/LightOnIO/status/1389579858754293761?s=20}},
to date the largest neural language model for French. Trained on the
CCnet corpus, we used scaling laws to inform our training setup,
resulting in an optimal use of our training budget. The evaluation of
our models on various end-tasks demonstrated first that the  CCnet corpus was a
better choice than the  Oscar French instance when used for
generation; second, they  showed that our models provide the same range of performance than their English
counterparts and established a new state of the art for summarization
of French on OrangeSum. PagnolXL and our smaller models are available
on \url{https://lair.lighton.ai/pagnol/}.

\section*{Acknowledgments}

This work was granted access to the HPC resources of IDRIS under the
allocation 2020-AD011012024 made by GENCI, enabling us to use the
\emph{Jean Zay} supercomputer. We thank Stéphane R\'{e}quena and the
support team for their valuable help. We also acknowledge Louis Martin
for helping us reproducing the CamemBERT settings for CCNet. Djamé
Seddah was partly funded by the French Research National Agency via
the ANR project ParSiTi (\mbox{ANR-16-CE33-0021}).

\bibliography{anthology,custom}
\bibliographystyle{acl_natbib}

\newpage

\appendix

\section{Pretraining details}
In this section we provide more details on the hyperparameters used to train PAGnol models. 

\paragraph{Models trained on OSCAR}
The Small and Medium trained on OSCAR use the Adam optimizer with $\varepsilon=1e-8$, with $2000$ warmup steps, and a cosine decay learning rate schedule over $325000$ steps. Additionally, we use grad clipping with a value of $1.0$, and set the dropout probability $p=0.1$. We do not use weight decay. The Small was trained on a single node with 8 NVIDIA V100, and the Medium on two such nodes. In both cases we use the distributed data parallel (DDP) paradigm, with mixed precision. We show differences in hyperparameters between the Small\textsubscript{OSCAR} and Medium\textsubscript{OSCAR} in Table \ref{tab:oscar_hyperparameters}
\begin{table}[]
    \centering
    \begin{tabular}{lcc}
    \toprule
        PAGnol \textsubscript{OSCAR} & S & M \\
         \midrule
        Shared Emb. & False & True \\
        $(\beta_1, \beta_2)$ & $(0.9, 0.95)$ & $(0.9, 0.999)$\\  
        LR\textsubscript{max-min} & 6e-4 -- 6e-5 & 3e-4 -- 3e-5\\
        Batch Size & 12 & 5\\
        \bottomrule
    \end{tabular}
    \caption{Hyperparameters differences between the Small and Medium training runs on OSCAR.}
    \label{tab:oscar_hyperparameters}
\end{table}
\begin{table}[]
    \centering
    \begin{tabular}{lc}
    \toprule
    PAGnol & GPU-hours \\
    \midrule
    S      & 480 \\
    M      & 1,920 \\
    L      & 6,720 \\
    XL     & 5,200 \\
    \midrule
    Total  & 14,320 \\
    \bottomrule
    \end{tabular}
    \caption{The Large model required more GPU-hours than expected, however several optimizations allowed us to perform the training of the XL model with a smaller budget.}
    \label{tab:gpu_hours}
\end{table}
\paragraph{Models trained on CC-100}
We use the Adam optimizer with $\varepsilon=1e-8$ and warmup and cosine learning rate decay schedule also for the S, M, L, and XL models trained on CC-100. We set dropout probability $p=0.1$ and no weight decay for all models. Small, Medium, and Large are trained with DDP, and use tied embeddings, gradient clipping at $1.0$, $(\beta_1, \beta_2)=(0.9, 0.999)$. These are trained on respectively 1, 2 and 4 nodes with 8 NVIDIA V100s. The XL model uses a smaller value for gradient clipping at $0.75$, a larger context size at $2048$, and different $(\beta_1, \beta_2)=(0.9, 0.95)$ for the Adam optimizer. This model is trained on 10 nodes with 4 NVIDIA V100s each. Further differences between the hyperparameters of the training runs are shown in Table \ref{tab:ccnet_hyperparameters}.
\begin{table*}[]
    \centering
    \begin{tabular}{lcccc}
    \toprule
        PAGnol \textsubscript{CC-100} & S & M & L & XL \\
         \midrule
         LR\textsubscript{max-min} & 5e-4--5e-5 & 3e-4--3e-5 & 2.5e-4--2.5e-5 & 2.5e-4--2.5e-5\\
         Warmup Steps & 5000 & 5000 & 5000 & 3600\\
         Decay Steps & 400,000 & 500,000 & 100,000 & 400,000\\
         Batch Size & 10 & 5 & 1 & 4\\
        \bottomrule
    \end{tabular}
    \caption{Hyperparameters differences between Small, Medium, Large and Extra-Large training runs on CC-100.}
    \label{tab:ccnet_hyperparameters}
\end{table*}
\paragraph{Budget in GPU-Hours}
We report in Table~\ref{tab:gpu_hours} the budget in GPU-hours needed to train each model, and the total budget.

\section{Samples from PAGnol XL}
We provide in Tables \ref{tab:random}, \ref{tab:oroscope}, \ref{tab:lemonde}, and \ref{tab:senscritique} some examples of completions given by PAGnol XL when prompted with different text.
\newpage
\begin{table*}
    \centering
    \begin{tabular}{p{0.2\linewidth}|p{0.6\linewidth}}
       Prompt  & Bonjour Madame, est-ce que \\
       Generation & \textit{le château de la Noue se trouve dans la région ? Est-ce que les gens s’y déplacent en calèche ? Je me rends dans ce coin du monde quelques jours au mois de mai, c’est mon anniversaire. Est-ce que vous pouvez me conseiller quelque chose ? Merci beaucoup. Françoise Vernin. }
    \end{tabular}
    \caption{Completion for a random French sentence.}
    \label{tab:random}
\end{table*}

\begin{table*}
    \centering
    \begin{tabular}{p{0.2\linewidth}|p{0.6\linewidth}}
       Prompt  & Produis un horoscope étant donné un signe astrologique.
----------
Bélier: Des nuages sur votre Ciel ! En couple, des moments de désaccord sont possibles. Vénus pourrait en effet compliquer vos relations avec votre conjoint...
----------
Taureau: Vénus donnera la priorité à vos amours ! En couple, les événements en cours vous seront favorables à court ou long terme, celui qui vous aime...
----------
Lion: Vos amours devraient vous combler. En couple, Jupiter vous offrira le cocooning harmonieux dont vous rêvez. Célibataire, inutile de papillonner:...
----------
Vierge: Ambiance astrale morose. En couple, vous serez probablement tenté de chercher un dérivatif à l'ennui qui naît de l'habitude et du train-train...
----------
Capricorne: \\
       Generation & \textit{Il est temps de changer de cap! En couple, Vénus s'opposera à Mars et Mercure. Ces deux planètes vous réjouiront par leur dynamisme. Mais, en tant que conjointe, vous serez plus fatiguée que jamais.}
    \end{tabular}
    \caption{Completion for an oroscope found in the magazine ELLE.}
    \label{tab:oroscope}
\end{table*}

\begin{table*}
    \centering
    \begin{tabular}{p{0.2\linewidth}|p{0.6\linewidth}}
       Prompt  & Quelque 180 000 nouveaux parents devraient en bénéficier d’ici au mois d’août 2022 : le gouvernement prévoit de distribuer des « bébés box » par l’intermédiaire des maternités à partir de février 2022. De quoi s’agit-il ? \\
       Generation & \textit{De boîtes contenant un kit composé de plusieurs couches, quelques vêtements et accessoires adaptés à l'âge et au sexe de l'enfant. Certains parents ont pu déjà recevoir ces kits grâce à une souscription mensuelle proposée par les enseignes de la grande distribution. L’offre a immédiatement rencontré un vif succès, des petits comme des grands, qui jouent à la fois un rôle de « maman » et de « papa » mais surtout avec ce budget...}
    \end{tabular}
    \caption{Completion for the beginning of a news article on the "« Bébé box » pour jeunes parents" from Le Monde website on the 14th of October 2021.}
    \label{tab:lemonde}
\end{table*}

\begin{table*}
    \centering
    \begin{tabular}{p{0.2\linewidth}|p{0.6\linewidth}}
       Prompt  & La septième réalisation du monumental Bong Joon Ho est fréquemment présentée comme une critique acerbe des inégalités qui minent la société coréenne. La lutte, d’abord larvée et de plus en plus patente, qui oppose deux familles aux extrêmes de l’échelle sociale semble étayer cette vision.\\
       Generation & \textit{Il ne s’agira donc pas de confronter les puissants à leurs limites mais plutôt de s’interroger sur le mensonge et la tromperie, les jeux et l’alcool, les privilèges et la corruption. Pour en arriver à cette conclusion, on n’a cependant pas la prétention de se demander comment une famille moyenne aisée pourrait être en mesure de maintenir les apparences dans un pays où il n’y a pas d’eau, ni de gaz, ni d’électricité. En tant que spectateur, nous suivons ce cadre tendu, ce huis clos. Puis, très vite, les événements s’ accélèrent, la mécanique bien huilée du thriller se met en marche pour un final imprévisible.}
    \end{tabular}
    \caption{Completion for the beginning of a critique on Parasite from the website \textit{Sens Critique}. Accessed on the 14th of October 2021.}
    \label{tab:senscritique}
\end{table*}

\end{document}